\documentclass{article}
\usepackage{spconf,amsmath,graphicx}
\usepackage[utf8]{inputenc}
\usepackage{times}
\usepackage{epsfig}
\usepackage{amssymb}
\usepackage[dvipsnames]{xcolor}
\usepackage{capt-of}
\usepackage{pifont}
\usepackage{tabularx}
\usepackage{setspace}

\definecolor{bblue}{rgb}{0.0,0.25,0.65}
\definecolor{ccol}{rgb}{0.2,0.2,0.2}
\usepackage[pagebackref=true,breaklinks=true,letterpaper=true,colorlinks=true,urlcolor=bblue,bookmarks=false,allcolors=black]{hyperref}

\newcolumntype{Y}{>{\centering\arraybackslash}X}
\newcolumntype{R}{>{\raggedleft\arraybackslash}X}
\newcolumntype{L}{>{\raggedright\arraybackslash}X}

\usepackage{expl3}
\ExplSyntaxOn
\newcommand\latinabbrev[1]{
	\peek_meaning:NTF . {
		#1\@}%
	{ \peek_catcode:NTF a {
			#1.\@ }%
		{#1.\@}}}
\ExplSyntaxOff
\def\eg{\latinabbrev{e.g}~}

\def\etc{\latinabbrev{etc}}
\def\ie{\latinabbrev{i.e}~}

\newcommand\customparagraph[1]{\vspace{0.6em}\noindent\textbf{#1}}

\title{Photorealistic Image Synthesis for Object Instance Detection}

\newcommand{\namesep}{\hspace{2ex}}
\makeatletter
\def\@name{
	\emph{Tomáš~Hodaň$^{1,2}$}\namesep
	\emph{Vibhav~Vineet$^{2}$}\namesep
	\emph{Ran~Gal$^{2}$}\namesep
	\emph{Emanuel~Shalev$^{2}$}\namesep
	\emph{Jon~Hanzelka$^{2}$}\\
	\emph{Treb~Connell$^{2}$}\namesep
	\emph{Pedro~Urbina$^{2}$}\namesep
	\emph{Sudipta~N.~Sinha$^{2}$}\namesep
	\emph{Brian~Guenter$^{2}$}\\
}
\makeatother
\address{
	$^{1}$FEE, Czech~Technical~University~in~Prague\namesep
	$^{2}$Microsoft~Research
}

\begin{document}
	
	\maketitle
	
	\begin{abstract}
		We present an approach to synthesize highly photorealistic images of 3D object models, which we use to train a convolutional neural network for detecting the objects in real images.
The proposed approach has three key ingredients:
(1)~3D object models are rendered in 3D models of complete scenes with realistic materials and lighting,
(2)~plausible geometric configuration of objects and cameras in a scene is generated using physics simulation, and
(3)~high photorealism of the synthesized images is achieved by physically based rendering.
When trained on images synthesized by the proposed approach, the Faster R-CNN object detector~\cite{ren2017faster} achieves a 24\% absolute improvement of mAP@.75IoU on Rutgers APC~\cite{rennie2016dataset} and 11\% on LineMod-Occluded~\cite{BrachmannKMGSR_eccv14} datasets, compared to a baseline where the training images are synthesized by rendering object models on top of random photographs.
This work is a step towards being able to effectively train object detectors without capturing or annotating any real images. A dataset of 600K synthetic images
with ground truth annotations for various computer vision tasks will be released~on the project website:
\texttt{\href{https://thodan.github.io/objectsynth/}{thodan.github.io/objectsynth}}.
	\end{abstract}

\begin{figure}[t!]
	\begin{center}
		\begin{tabular}{ @{}c@{ } @{}c@{ } }
			\includegraphics[width=0.49\linewidth]{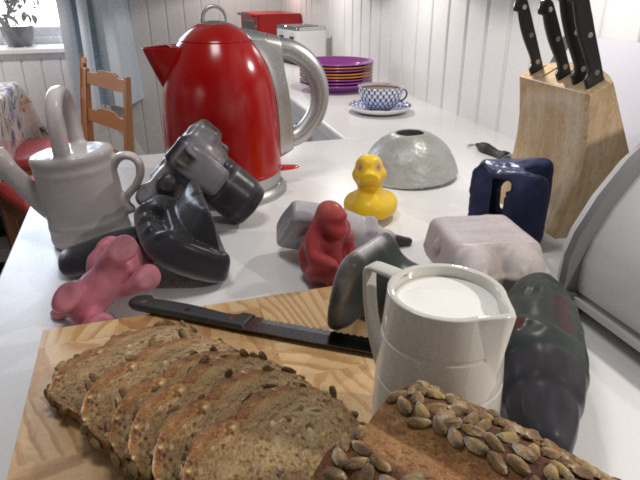} &
			\includegraphics[width=0.49\linewidth]{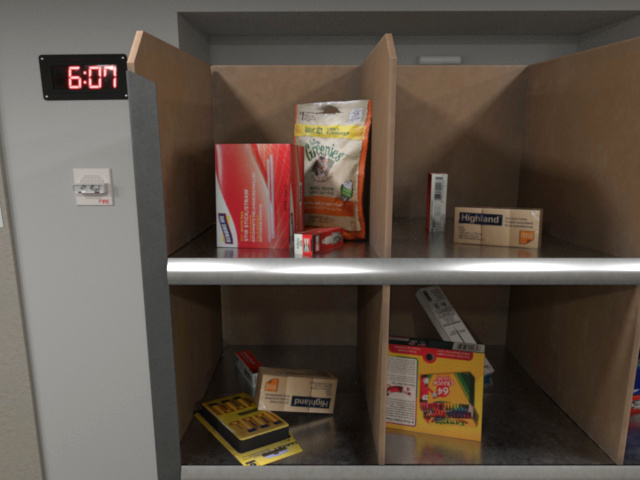} \\
			\multicolumn{2}{c}{\small{Photorealistic images synthesized by the proposed approach}} \vspace{1.5ex} \\
			\includegraphics[width=0.49\linewidth]{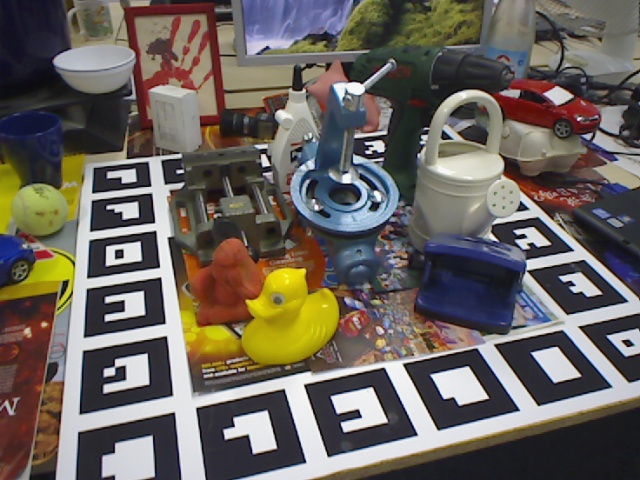} &
			\includegraphics[width=0.49\linewidth]{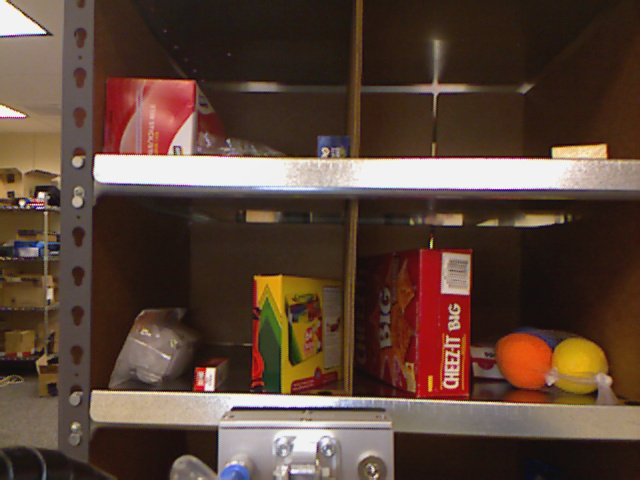} \\
			\multicolumn{2}{c}{\small{Real images from LineMod~\cite{hinterstoisser2012accv} and Rutgers APC~\cite{rennie2016dataset} datasets}} \\
		\end{tabular}
		\caption{\label{fig:teaser}
			Faster R-CNN object detector~\cite{ren2017faster} achieves 11--24\% higher mAP@.75IoU on real test images when trained on photorealistic synthetic images than when trained on images of objects rendered on top of random photographs.
			\vspace{-1ex}
		}
	\end{center}
\end{figure}

\section{Introduction}

Object instance detection is a
computer vision task which involves recognizing specific objects in an image and estimating their
2D bounding boxes.
Convolutional neural networks (CNN's) have become the standard approach for tackling this task. However,
training CNN models requires large amounts of real annotated images which are
expensive to acquire.

Computer graphics
has been used to synthesize training images for various computer vision tasks. This approach scales well as only minimal human effort, which may include 3D modeling, is required.
Nevertheless, despite training CNN's on massive datasets of diverse synthetic images,
a~large drop of performance has been observed when models trained only on synthetic images were tested on real images~\cite{RichterVRK_eccv16, SuQLG15, Rozantsev2018}.
The domain gap between the synthetic and real images can be reduced by domain adaptation techniques that aim to learn domain invariant representations or to transfer trained models from one domain to another~\cite{Csurka_acvpr17}.
A different line of work, presumably complementary to
domain adaptation,
has recently tried to reduce the domain gap by synthesizing training images with a higher degree of visual realism. The use of physically based rendering has been considered with this motivation and shown promising results~\cite{LiS_eccv18,ZhangSYSLJF_cvpr17}.

Physically based rendering techniques, \eg~Arnold~\cite{georgiev2018arnold}, accurately simulate the flow of light energy in the scene by ray tracing. This naturally accounts for complex illumination effects such as scattering, refraction and reflection, including diffuse and specular interreflection between the objects and the scene and between the objects themselves.
The rendered images are very realistic and often difficult to differentiate from real photographs~\cite{pharr2016physically}.
Rendering techniques based on rasterization, \eg~OpenGL~\cite{shreiner2009opengl}, can approximate the complex effects in an ad hoc way through custom shaders, but the approximations cause physically incorrect artifacts that are difficult to eliminate~\cite{marschner2015fundamentals}.
Physically based rendering has been historically noticeably slower than rasterization, however, the recently introduced Nvidia RTX ray tracing GPU promises a substantial reduction of the rendering time.

In this work, we investigate the use of highly photorealistic synthetic images for training Faster R-CNN, a CNN-based object detector~\cite{ren2017faster}.
To synthesize the images, we present an approach with three key ingredients.
First, 3D models of objects are not rendered in isolation but inside 3D models of complete scenes.
For this purpose, we have created models of six indoor scenes with realistic materials and lighting.
Second, plausible geometric configuration of objects and cameras in a scene is generated using physics simulation.
Finally, a high degree of visual realism
is achieved by physically based rendering (see Fig.~\ref{fig:teaser} and the supplementary material).

The experiments show that Faster R-CNN trained on the photorealistic synthetic images achieves a 24\% absolute improvement of mAP@.75IoU on real test images from Rutgers APC~\cite{rennie2016dataset}, and 11\% on real test images from LineMod-Occluded~\cite{BrachmannKMGSR_eccv14,hinterstoisser2012accv}. The improvement is relative to a baseline where the training images are synthesized by rendering object models on top of random photographs -- similar images are commonly used for training methods for
tasks such as object instance detection~\cite{dwibedi2017cut}, object instance segmentation~\cite{hinterstoisser2017pre}, and 6D object pose estimation~\cite{kehl2017ssd,rad2017bb8}.

	\section{Related Work} \label{sec:related_work}

Synthetic images have been used for benchmarking and training models for various computer vision tasks.
Here we review
approaches to generate synthetic images and approaches to reduce the gap between the synthetic and real domains.

\customparagraph{Rendering Objects.}
Su et al.~\cite{SuQLG15} synthesized images of 3D object models
for viewpoint estimation, Hinterstoisser et al.~\cite{hinterstoisser2017pre} for object instance detection and segmentation, and Dosovitskiy et al.~\cite{dosovitskiy2015flownet}
for optical flow estimation. They used a fixed
OpenGL
pipeline and pasted the rendered pixels over
randomly selected real photographs.
Rad et al.~\cite{rad2017bb8}, Tekin et al.~\cite{TekinSF_corr17} and Dwibedi et al.~\cite{dwibedi2017cut} similarly pasted segments of objects from real images on other real images for
object detection and pose estimation.
Dvornik et al.~\cite{dvornik2018} showed the importance of selecting suitable background images.
While these approaches are easy to implement, the resulting images are not realistic, objects often have inconsistent shading with respect to the background scene, interreflections and shadows are missing and the object pose and context are usually not natural.
Attias et al. \cite{Movshovitz-Attias_corr16} rendered photorealistic images of 3D car models placed within 3D scene models and showed the benefit over naive rendering methods, whereas Tremblay et al.~\cite{tremblay2018deep} rendered objects in physically plausible poses in diverse scenes,
but did not use physically based rendering.

\customparagraph{Rendering Scenes.}
Another line of work explored rendering of complete scenes and generating corresponding ground truth maps.
Richter et al. \cite{RichterHK_iccv17, RichterVRK_eccv16} leveraged existing commercial game engines to acquire ground truth for
several tasks. However, such game engines cannot be customized to insert new 3D object models.
Synthia~\cite{RosSMVL_cvpr16} and Virtual KITTI~\cite{gaidon2016virtual} datasets were generated using virtual cities modeled from scratch. Handa et al. \cite{HandaPBSC_corr15a} and Zhang \cite{ZhangSYSLJF_cvpr17} modeled 3D scenes for semantic scene understanding.
Finally, SceneNet~\cite{mccormac2017scenenet} was created by synthesizing RGB-D video frames from simulated cameras moving realistically within static scenes.

\customparagraph{Domain Adaptation.}
Popular approaches to bridge the gap between the synthetic and real domains include re-training the model in the real domain, learning domain invariant features, or learning a mapping between the two domains~\cite{tobin2017domain,Csurka_acvpr17}.
In contrast, domain randomization methods reduce the gap by randomizing the rendering parameters and have been used in object localization and pose estimation~\cite{tobin2017domain, TremblayPABJATCBB_corr18, SundermeyerMDBT_eccv18}.

\customparagraph{PBR based approaches.}
Physically based rendering (PBR) techniques
can synthesize images with a high degree of visual realism which promises to reduce the domain gap.
Li and Snavely \cite{LiS_eccv18} used PBR
images to train models for intrinsic image decomposition
and Zhang et al. \cite{ZhangSYSLJF_cvpr17}
for semantic segmentation, normal estimation and boundary detection.
However, they focus on scene understanding not object understanding tasks.
Wood et al.~\cite{wood2015rendering} used PBR images of eyes for training gaze estimation models. Other ways to generate photorealistic images
have been also proposed~\cite{wood2016learning, shrivastava2017learning}.

\section{Proposed Approach}
\label{sec:synthesis}

\begin{figure}[t!]
	\begin{center}
		\includegraphics[width=1.0\columnwidth]{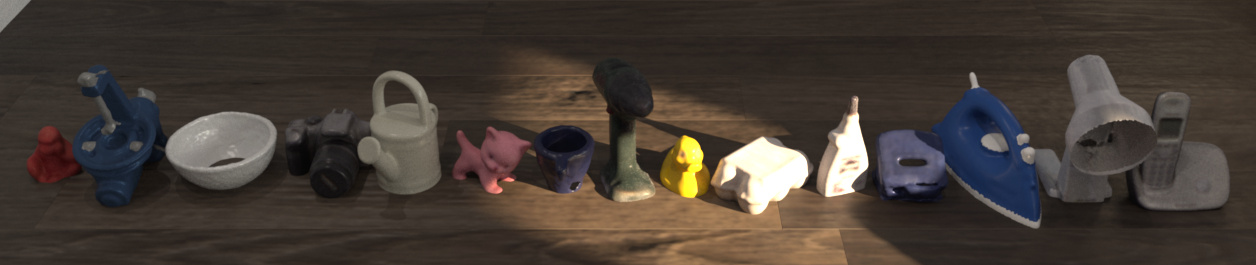} \\
		\vspace{0.3ex}
		\includegraphics[width=1.0\columnwidth]{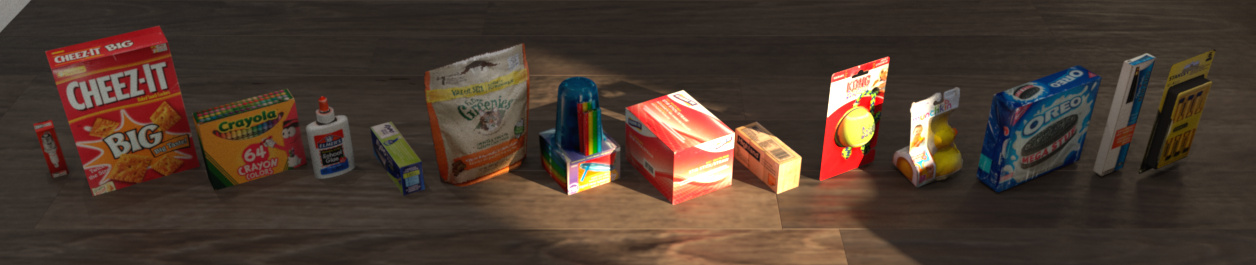} \\
		\vspace{0.5ex}
		\begin{tabular}{ @{}c@{ } @{}c@{ } @{}c@{ } }
			\small{Scene 1} & \small{Scene 2} & \small{Scene 3} \\
			\includegraphics[width=0.3265\columnwidth]{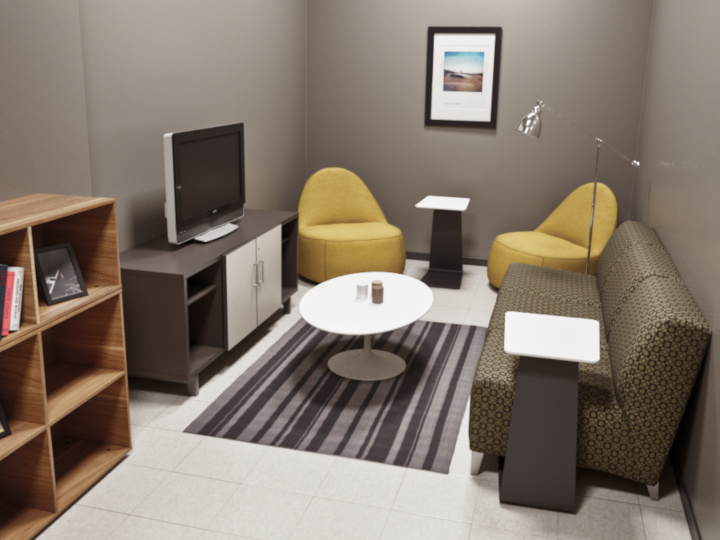} &
			\includegraphics[width=0.3265\columnwidth]{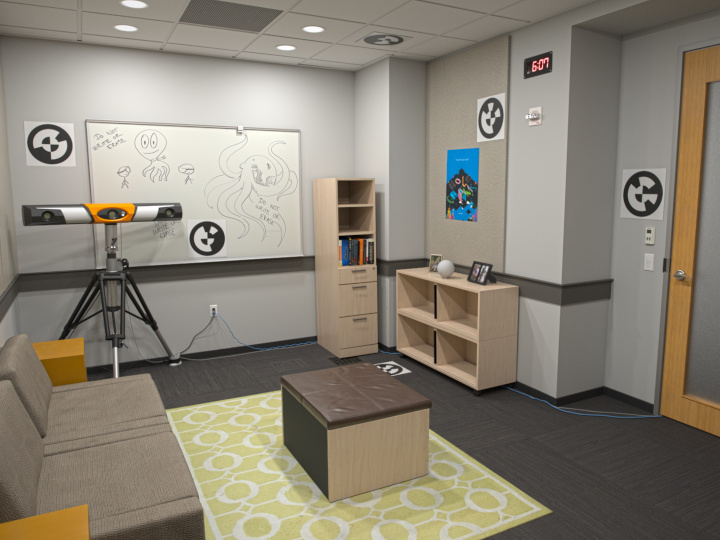} &
			\includegraphics[width=0.3265\columnwidth]{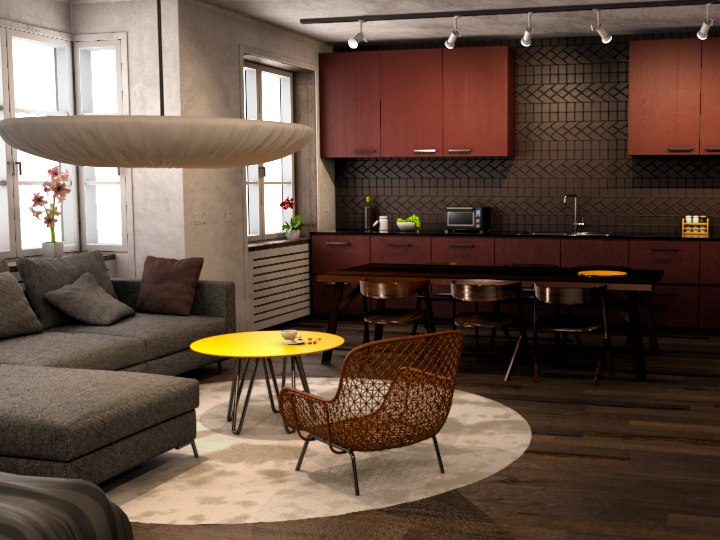} \\
			\small{Scene 4} & \small{Scene 5} & \small{Scene 6} \\
			\includegraphics[width=0.3265\columnwidth]{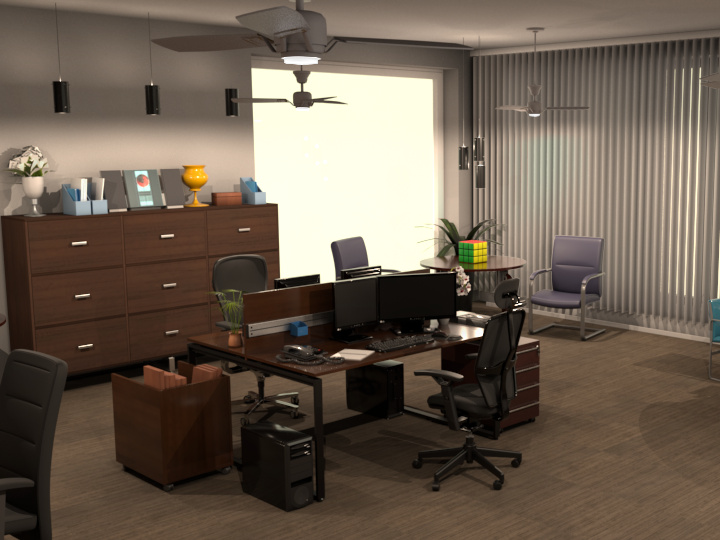} &
			\includegraphics[width=0.3265\columnwidth]{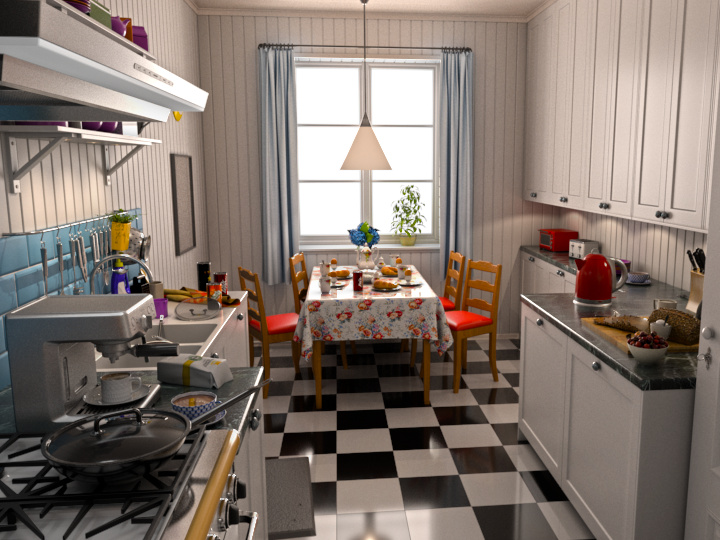} &
			\includegraphics[width=0.3265\columnwidth]{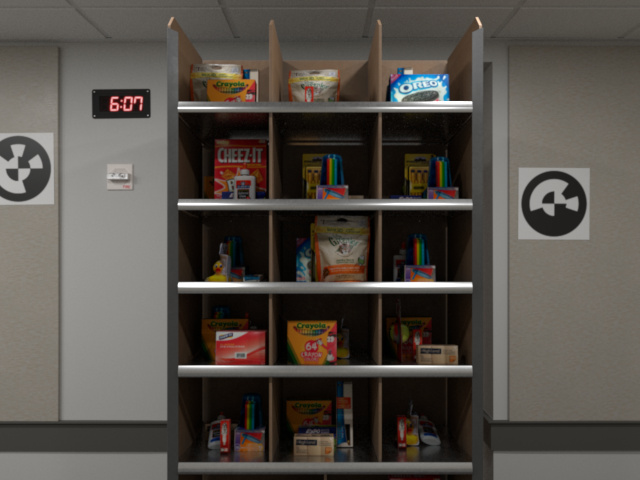}
		\end{tabular}
		\caption{\label{fig:models} 3D object models from LineMod~\cite{hinterstoisser2012accv} (first row) and Rutgers APC~\cite{rennie2016dataset} (second row) were rendered in six scenes.
		}
	\end{center}
\end{figure}

To achieve a high degree of visual realism in computer generated imagery, one needs to focus on (1) modeling the scene to a high level of detail in terms of geometry, textures and materials,
and (2) simulating the lighting, including soft shadows, reflections, refractions and indirect light bounces~\cite{georgiev2018arnold}.
This section describes the proposed approach to synthesize highly photorealistic images of objects in indoor scenes, which includes modeling of the objects and the scenes (Fig.~\ref{fig:models}), arranging the object models in the scene models, generating camera poses, and rendering images of the object arrangements.

\subsection{Scene and Object Modeling}

\customparagraph{3D Object Models.}
We worked with 3D models of 15 objects from LineMod (LM)~\cite{hinterstoisser2012accv} and 14 objects from Rutgers APC (RU-APC)~\cite{rennie2016dataset}. We used the refined models from BOP~\cite{hodan2018bop}, provided as colored meshes with surface normals, and manually assigned them material properties.
A Lambertian material was used for the RU-APC models as the objects are made mostly of cardboard.
The LM models were assigned material properties which match their appearance in real images.
The \emph{specular}, \emph{metallic} and \emph{roughness} parameters of the Arnold renderer~\cite{georgiev2018arnold} were used to control the material properties.

\customparagraph{3D Scene Models.}
The object models were arranged and rendered within 3D models of six furnished scenes.
Scenes 1--5 represent work and household environments and include fine details and typical objects, \eg the kitchen scene (Scene~5) contains dishes in a sink
or a bowl of cherries. Scene~6 contains a shelf from the Amazon Picking Challenge 2015~\cite{yu2016summary}.

The scene models were created using standard 3D tools, primarily Autodesk Maya.
Scenes 1 and 2 are reconstructions of real-world environments obtained using LiDAR and photogrammetry 3D scans which served as a
guide for an artist.
Materials were recreated using photographic reference, PBR material scanning~\cite{muravision}, and color swatch samples~\cite{nixsensor}.
Scenes 3--5 were purchased online~\cite{evermotion}, their geometry and materials were refined, and clutter and chaos was added to mimic a real environment.
A 3D geometry model of the shelf in Scene~6 was provided in the Amazon Picking Challenge 2015~\cite{yu2016summary}.
Reference imagery of the shelf was used to create textures and materials that
match its appearance.

Exterior light was modeled with Arnold Physical Sky~\cite{georgiev2018arnold}
which can accurately depict atmospheric effects and time-of-day variation. Interior lights
were modeled with standard light sources such as area and point lights.

\subsection{Scene and Object Composition} \label{sec:scene_composition}

\customparagraph{Stages for Objects.} In each scene, we manually selected multiple stages to arrange the objects on. A stage is defined by a polygon and is typically located on tables, chairs and other places with distinct structure, texture or illumination. Placing objects on such locations maximizes the diversity of the rendered images.
One stage per shelf bin was added in Scene~6.

\customparagraph{Object Arrangements.}
An arrangement of a set of objects was generated in two steps: (1) poses of the 3D object models were instantiated above one of the stages, and (2) physically plausible arrangements were reached using physics simulation where the objects fell on the stages under gravity and underwent mutual collisions.
The poses were initialized using FLARE~\cite{gal2014flare}, a rule-based system for generation of object layouts for
AR applications.
The initial height above the stage was randomly sampled from 5 to 50cm. For LM objects, we staged one instance per object model and initialized it with the canonical orientation, \ie the cup was up-right, the cat was standing on her legs, \etc. For RU-APC objects, we staged up to five instances per object model and initialized their orientation randomly.
Physics was simulated using NVIDIA PhysX.

\customparagraph{Camera Positioning.}
Multiple cameras
were positioned around each object arrangement.
Instead of fitting all the objects within the camera frustum, we point the camera at a randomly selected object. This allows for a better control of scale of the rendered objects.
The azimuth and elevation of the camera was generated randomly and the distance of the camera from the focused object was sampled from a specified range.
Before rendering the RGB image, which is computationally expensive, we first rendered a mask of the focused object and a mask of its visible part. We then calculated the visible fraction of the focused object and rendered the full RGB image only if the object was at least 30\% visible.

\subsection{Physically Based Rendering (PBR)}

Three images were rendered from each camera using the Arnold physically based renderer~\cite{georgiev2018arnold} at \emph{low}, \emph{medium} and \emph{high} quality settings -- the mapping between the quality settings and Arnold parameters can be found in the supplementary material.
Rendering was done on 16-core Intel Xeon 2.3GHz processors with 112GB RAM. GPUs were not used. The average rendering time was 15s in low, 120s in medium, and 720s in high quality settings.
We used a CPU cluster with 400 nodes which allowed us to render 2.3M images in low, 288K in medium, or 48K in high quality within a day.

We rendered 1.9M object instances in 1K arrangements in the six scenes, seen from 200K cameras. With the three quality settings, we obtained a total of 600K 
VGA resolution images.
LM objects were rendered in Scenes 1--5 and RU-APC objects in Scenes 3 and~6. 
Each object instance is annotated with a 2D bounding box, a segmentation mask and a 6D pose.

\section{Experiments} \label{sec:experiments}

The experiments evaluate the effectiveness of PBR images for training the Faster R-CNN object detector~\cite{ren2017faster}.
Specifically, the experiments focus on three aspects:
(1)~importance of the PBR images over the commonly used images of objects rendered on top of random photographs, (2)~importance of the high PBR quality,
and (3)~importance of scene context.

\customparagraph{Datasets.}
The experiments were conducted on two datasets, LineMod-Occluded (LM-O)~\cite{BrachmannKMGSR_eccv14,hinterstoisser2012accv} and Rutgers APC
(RU-APC)~\cite{rennie2016dataset}. We used their reduced versions from BOP~\cite{hodan2018bop}.
The datasets include 3D object models and real test RGB-D images of VGA resolution (only RGB channels were used). The images are annotated with ground-truth 6D object poses from which we calculated 2D bounding boxes used for evaluation of the 2D object detection task.
LM-O contains 200 images
with ground truth annotations for 8 LM objects captured with various levels of occlusion.
RU-APC contains 14 object models
and 1380 images which show the objects in a cluttered warehouse shelf.
Example test images are in Fig.~\ref{fig:teaser}.

\customparagraph{Baseline Training Images (BL).}
We followed the synthetic data generation pipeline presented in~\cite{hinterstoisser2017pre}
and used OpenGL to render 3D object models on top of randomly selected real photographs
pulled from NYU Depth Dataset V2~\cite{silberman2012indoor}.
For a more direct comparison, the BL images were rendered from the same cameras
as the PBR images, \ie the objects appear in the same poses in both types of images.
Generation of one BL image took 3s on average. Examples are in the supplement.

\customparagraph{Object Instance Detection.}
We experimented with two underlying network architectures of Faster R-CNN:
ResNet-101~\cite{he2016deep} and Inception-ResNet-v2~\cite{szegedy2017inception}.
The networks were pre-trained on Microsoft COCO~\cite{lin2014microsoft} and fine-tuned on synthetic images for 100K iterations.
The learning rate was set to 0.001 and multiplied by 0.96 every 1K iterations.
To virtually increase diversity of the training set, the images were augmented by randomly adjusting brightness, contrast, hue, and saturation, and by applying random Gaussian noise and blur.
Although the presented results were obtained with this data augmentation, we found its effect negligible.
Implementation from Tensorflow Object Detection API~\cite{huang2017speed} was used.

\customparagraph{Evaluation Metric.}
The performance
was measured by mAP@.75IoU, \ie the mean average precision with a strict IoU threshold of 0.75~\cite{lin2014microsoft}.
For each test image, detections of object classes annotated in the image
were considered.

\subsection{Importance of PBR Images for Training}
On RU-APC, Faster R-CNN with Inception-ResNet-v2 trained on high quality PBR images achieves a significant 24$\%$ absolute improvement of mAP@.75IoU over the same model trained on BL images (Tab.~\ref{tab:det_scores}, PBR-h vs. BL).
It is noteworthy that PBR images yield almost 35$\%$ or higher absolute improvement on five object classes, and overall achieve a better performance on 12 out of 14 object classes (see the supplement for the per-class scores).
This is achieved when the objects are rendered in Scene 6.
When the objects are rendered in Scene 3 (PBR-ho vs. BL), we still observe a large improvement of 11$\%$
(scene context is discussed in Sec.~\ref{sec:context}).
Improvements, although not so dramatic, can be observed also
with ResNet-101.
On LM-O, PBR images win by almost 11$\%$,
with a large improvement on 7 out of 8 object classes.

\begin{table}[t!]
	\begin{center}
		\footnotesize
		\begin{tabularx}{\columnwidth}{|c|c||Y|Y|Y|Y|}
			\hline
			{\scriptsize Dataset} & {\scriptsize Architecture} & {\scriptsize PBR-h} & {\scriptsize PBR-l} & {\scriptsize PBR-ho} & {\scriptsize BL} \\
			\hline\hline
			{\scriptsize LM-O} & {\scriptsize Inc.-ResNet-v2} & 55.9 & 49.8 & -- & 44.7 \\
			& {\scriptsize ResNet-101} & 49.9 & 44.6 & -- & 45.1 \\
			\hline
			{\scriptsize RU-APC} & {\scriptsize Inc.-ResNet-v2} & 71.9 & 72.9 & 58.7 & 48.0 \\	
			& {\scriptsize ResNet-101} & 68.4 & 65.1 & 51.6 & 52.7 \\								
			\hline
		\end{tabularx}
		\caption{\label{tab:det_scores} Performance (mAP@.75IoU) of Faster R-CNN trained on high and low quality PBR images (PBR-h, PBR-l), high quality PBR images of out-of-context objects (PBR-ho), and images of objects on top of random photographs~(BL).
		\vspace{-1ex}
		}
	\end{center}
\end{table}

\subsection{Importance of PBR Quality}

On LM-O, we observe that Faster R-CNN with Inception-ResNet-v2 trained on high quality PBR images achieves an improvement of almost 6$\%$ over low quality PBR images (Tab.~\ref{tab:det_scores}, PBR-h vs. PBR-l).
This suggests that a higher PBR quality helps.
We do not observe a similar improvement on RU-APC when training on PBR images rendered in Scene~6.
The illumination in this scene is simpler, there is no incoming outdoor light and the materials are mainly Lambertian. There are therefore no complex reflections and the low quality PBR images from this scene are cleaner than, e.g., when rendered in Scenes 3--5. This suggests that the low quality is sufficient for scenes with simpler illumination and materials.
Example images of the two qualities are in the supplement.

\subsection{Importance of Scene Context} \label{sec:context}

Finally, we analyze the importance of accurately modeling the scene context.
We rendered RU-APC objects in two setups: 1)~\emph{in-context} in
Scene~6, and 2)~\emph{out-of-context} in Scene~3 (examples are in the supplement).
Following the taxonomy of contextual information from~\cite{divvala2009empirical},
the in-context setup faithfully model the gist, geometric, semantic, and illumination contextual aspects of the test scene. The out-of-context setup exhibit discrepancies in all of these aspects.
Training images of in-context objects yield an absolute improvement of 13$\%$ with Inception-ResNet-v2 and 16$\%$ with ResNet-101 over the images of out-of-context objects.
This shows the benefit of accurately modeling context of the test scene.

	\section{Conclusion} \label{sec:conclusion}

We have proposed an approach to synthesize highly photorealistic images of 3D object models
and demonstrated their benefit for training the Faster R-CNN object detector.
In the future, we will explore the use of photorealistic rendering for training models for other vision tasks.
A dataset of 600K photorealistic images will be released on the project website.

\vspace{2ex}
{\linespread{1.0}\selectfont{}
\begin{footnotesize}

\noindent We acknowledge K. Bekris and C. Mitash for help with the RU-APC dataset. This work was supported by V3C -- Visual Computing Competence Center (program TE01020415 funded by Technology Agency of the Czech Republic), Software Competence Center Hagenberg, and Research Center for Informatics (project CZ.02.1.01/0.0/0.0/16\_019/0000765 funded by OP VVV).

\end{footnotesize}
}
\normalsize

	{\small
		\bibliographystyle{IEEEbib}
		\bibliography{ref}
	}
	
\end{document}


\maketitle

\newpage

\begin{figure}[h!]
	\begin{center}
		\begin{tabular}{ @{}c@{ } @{}c@{ } }
			\includegraphics[width=0.495\columnwidth]{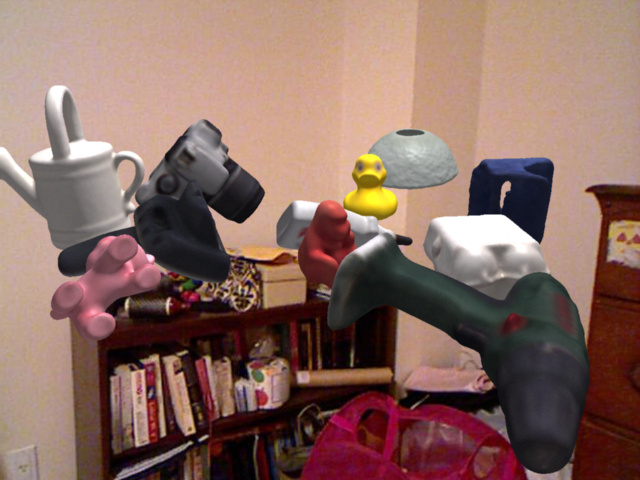} &
			\includegraphics[width=0.495\columnwidth]{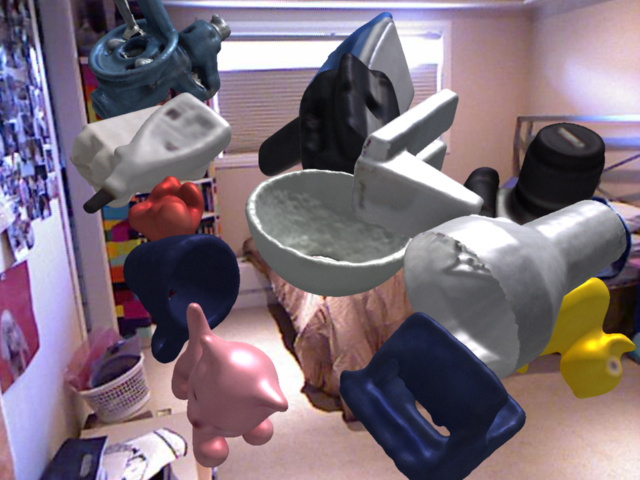} \\
		\end{tabular}
		\vspace{-2ex}
		\caption{\label{fig:random_bg}
			Examples of baseline training images generated by rendering 3D object models on top of random photographs.
		}
	\end{center}

	\begin{center}
		\begin{tabular}{ @{}c@{ } @{}c@{ } }
			\includegraphics[width=0.495\columnwidth]{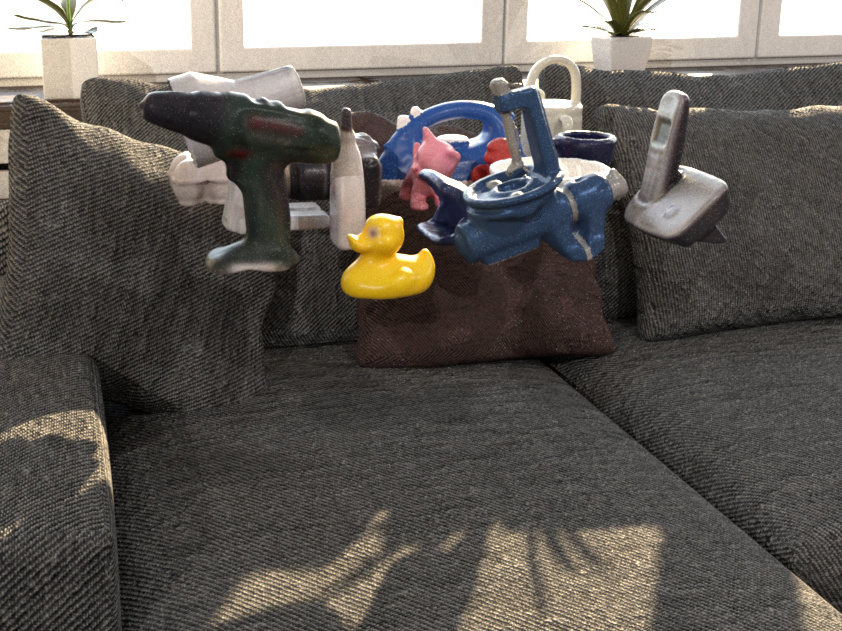} &
			\includegraphics[width=0.495\columnwidth]{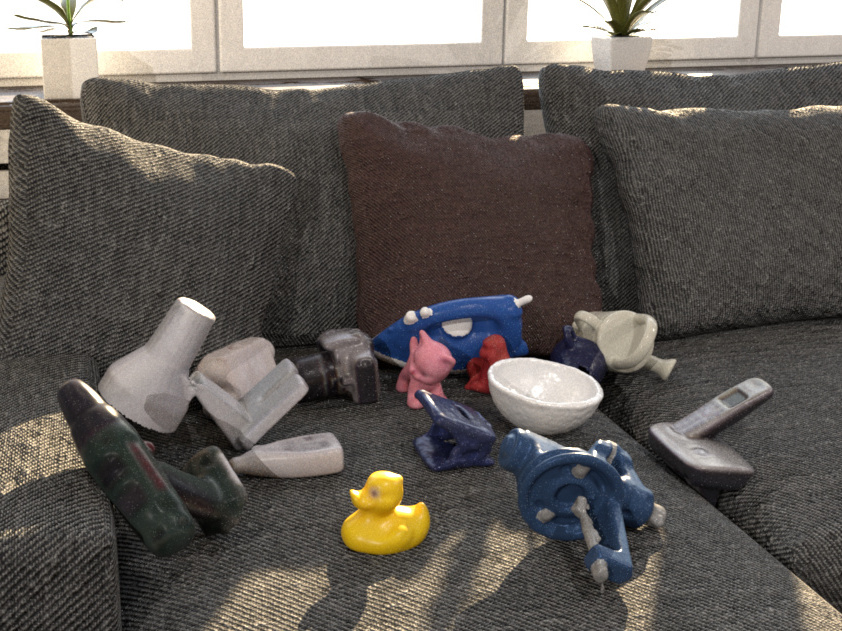} \\
		\end{tabular}
		\vspace{-2ex}
		\caption{\label{fig:arrangement} Initial object poses generated using FLARE~\cite{gal2014flare} (left), and final poses calculated by NVIDIA PhysX (right).}
	\end{center}

	\begin{center}
		\begin{tabular}{ @{}c@{ } @{}c@{ } }
			\includegraphics[width=0.495\columnwidth]{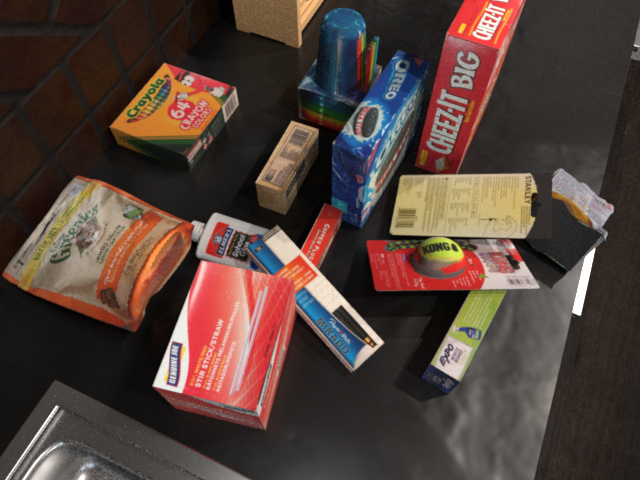} &
			\includegraphics[width=0.495\columnwidth]{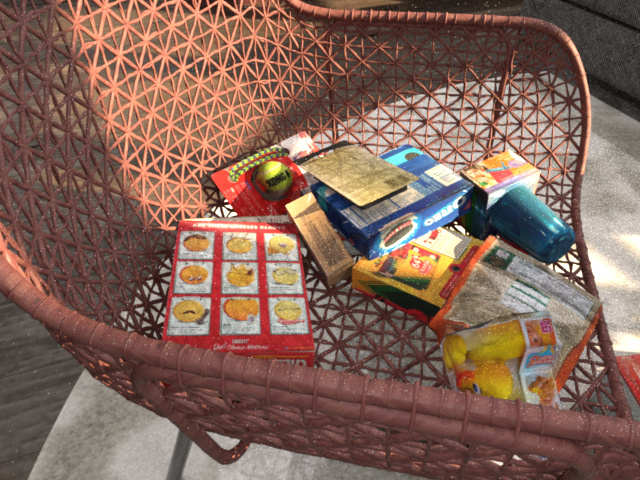} \\
		\end{tabular}
		\vspace{-2ex}
		\caption{\label{fig:apc_out_of_context}
		Training images of out-of-context RU-APC objects.
		}
	\end{center}	
\end{figure}

\begin{figure*}[h!]
	\begin{center}
		\begin{tabular}{ @{}c@{ } @{}c@{ } @{}c@{ } @{}c@{ } }
			\includegraphics[width=0.245\linewidth]{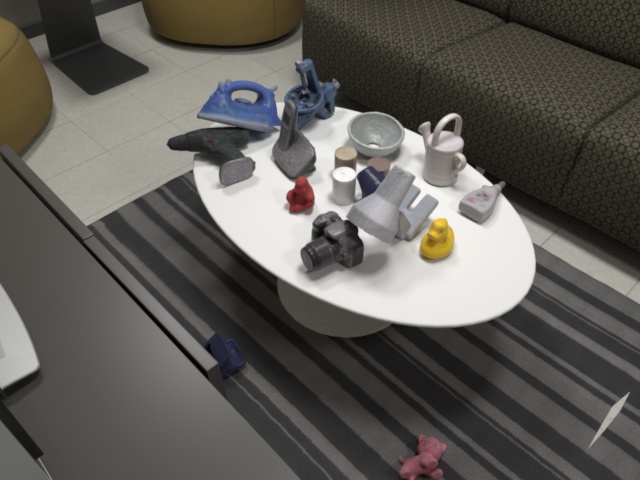} &
			\includegraphics[width=0.245\linewidth]{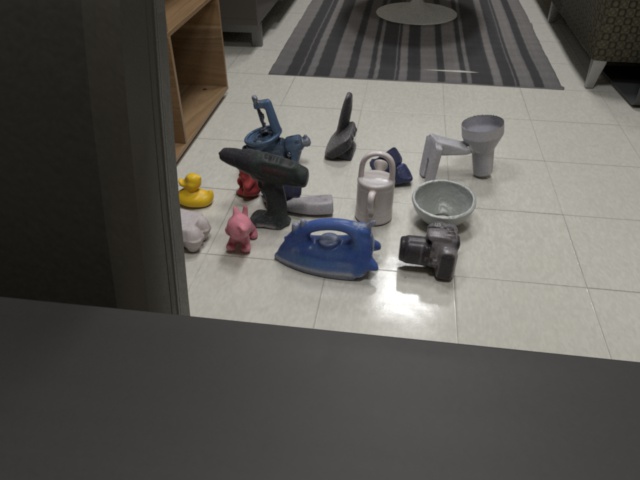} &
			\includegraphics[width=0.245\linewidth]{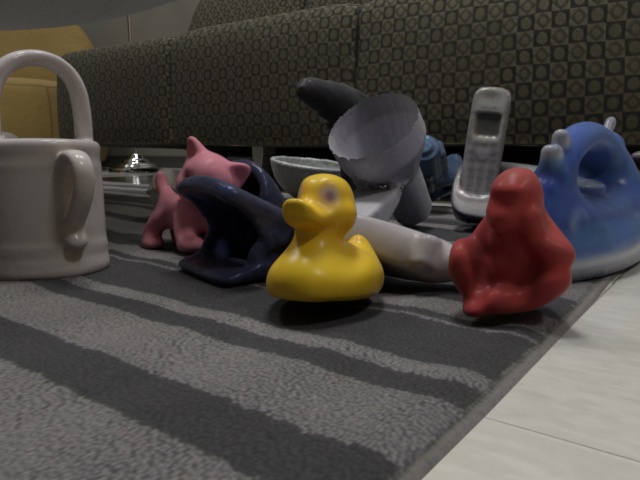} &
			\includegraphics[width=0.245\linewidth]{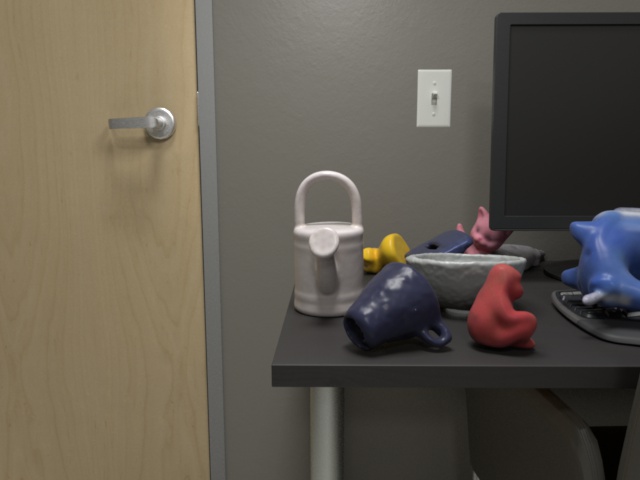} \\
			\includegraphics[width=0.245\linewidth]{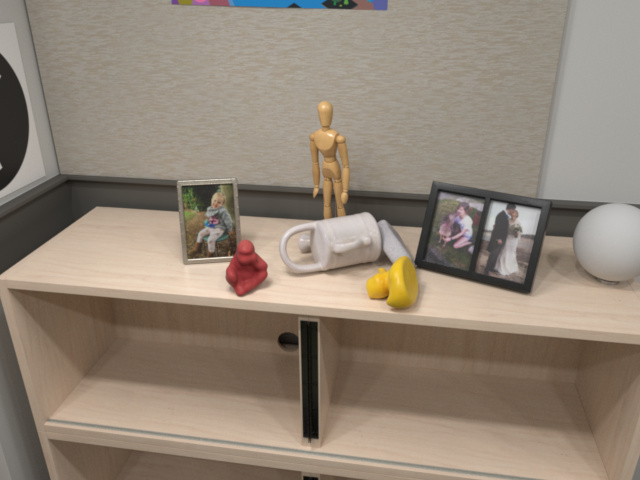} &
			\includegraphics[width=0.245\linewidth]{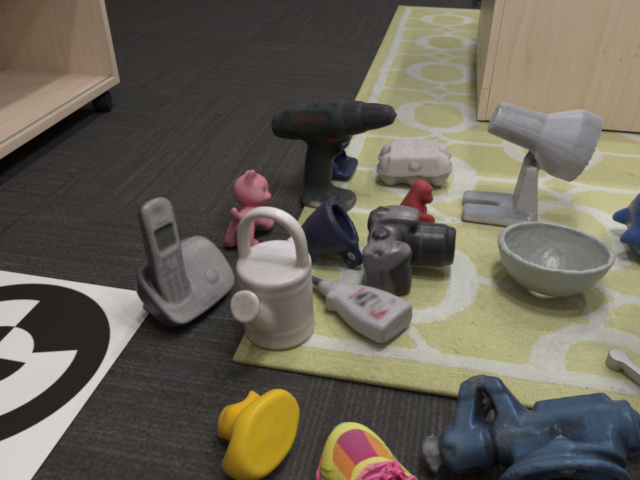} &
			\includegraphics[width=0.245\linewidth]{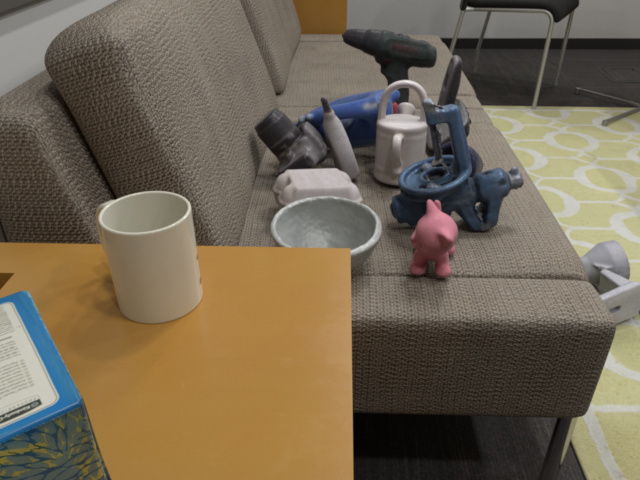} &
			\includegraphics[width=0.245\linewidth]{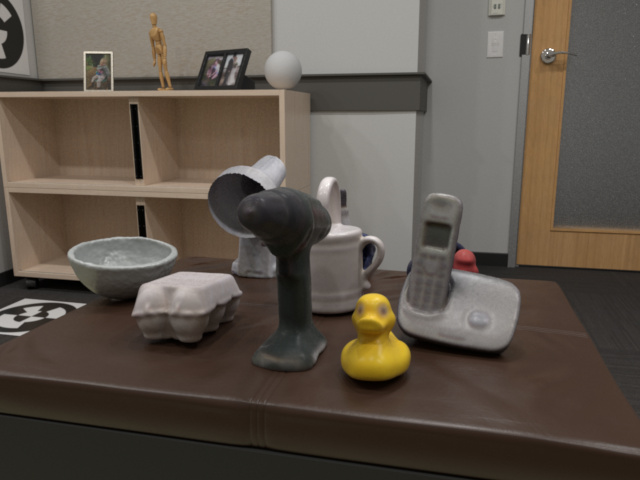} \\
			\includegraphics[width=0.245\linewidth]{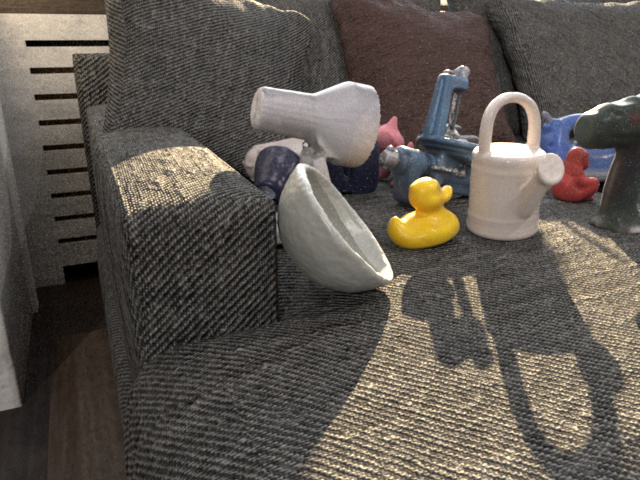} &
			\includegraphics[width=0.245\linewidth]{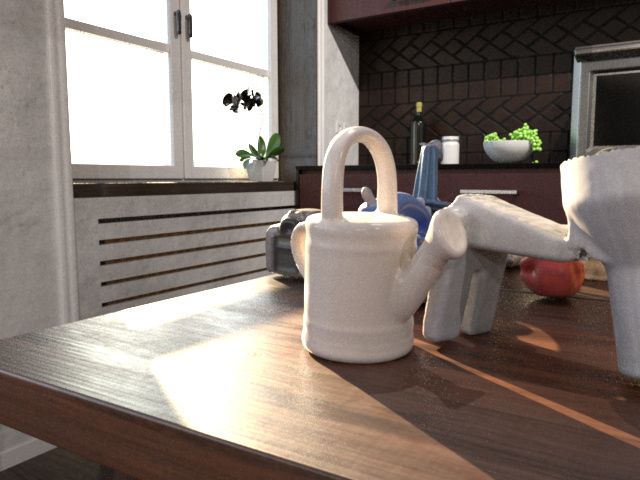} &
			\includegraphics[width=0.245\linewidth]{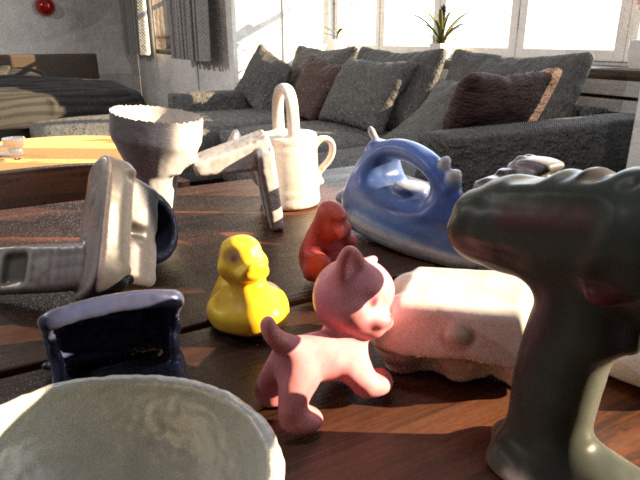} &
			\includegraphics[width=0.245\linewidth]{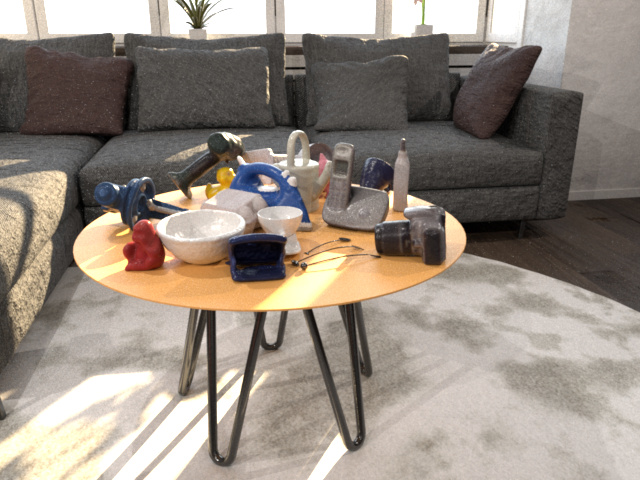} \\
			\includegraphics[width=0.245\linewidth]{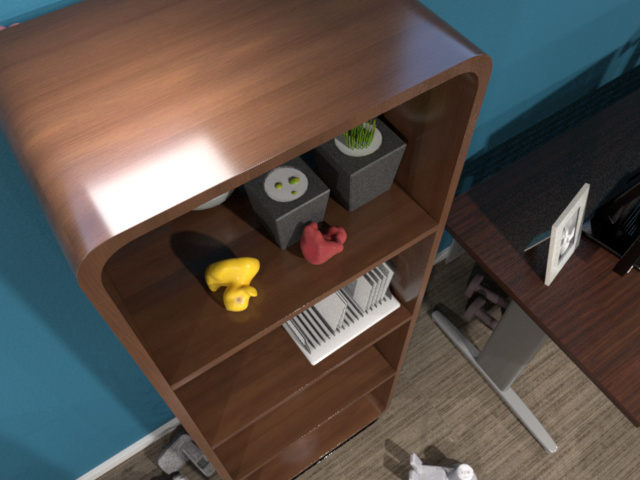} &
			\includegraphics[width=0.245\linewidth]{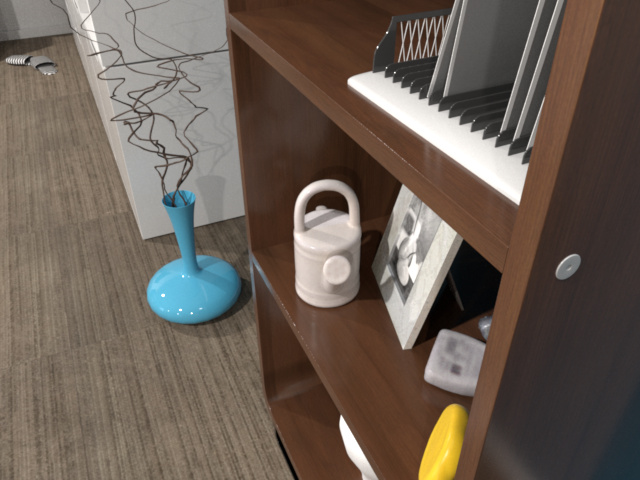} &
			\includegraphics[width=0.245\linewidth]{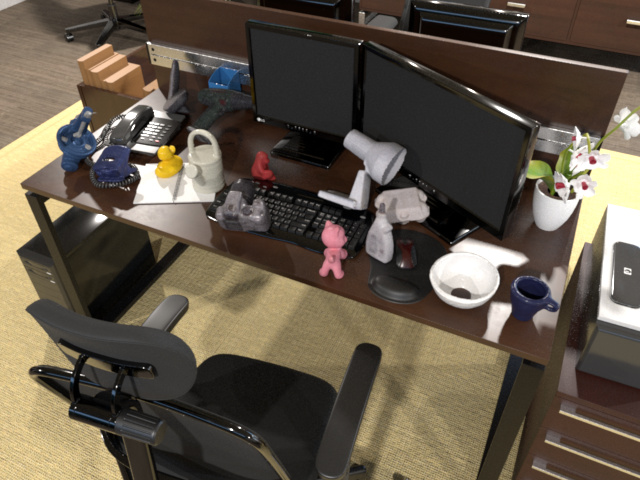} &
			\includegraphics[width=0.245\linewidth]{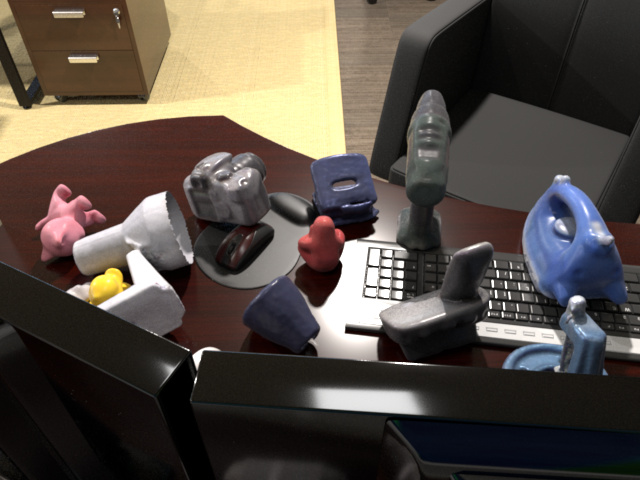} \\
			\includegraphics[width=0.245\linewidth]{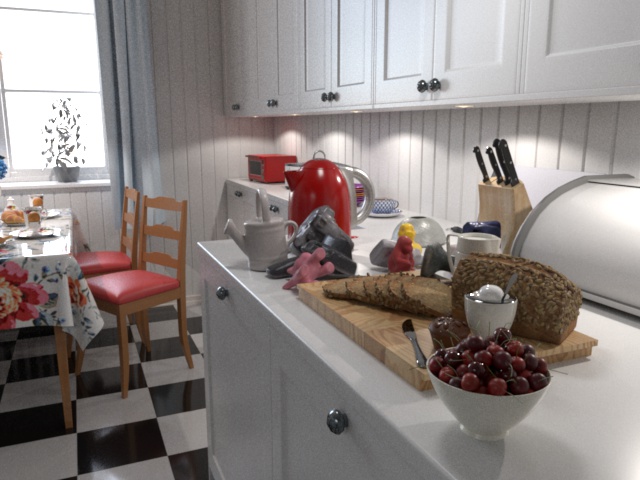} &
			\includegraphics[width=0.245\linewidth]{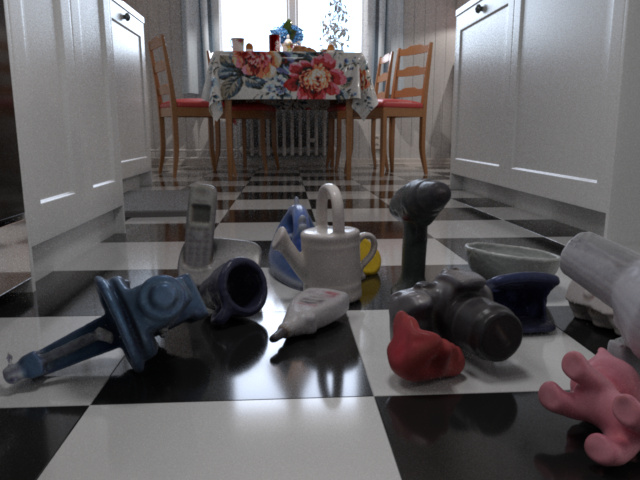} &
			\includegraphics[width=0.245\linewidth]{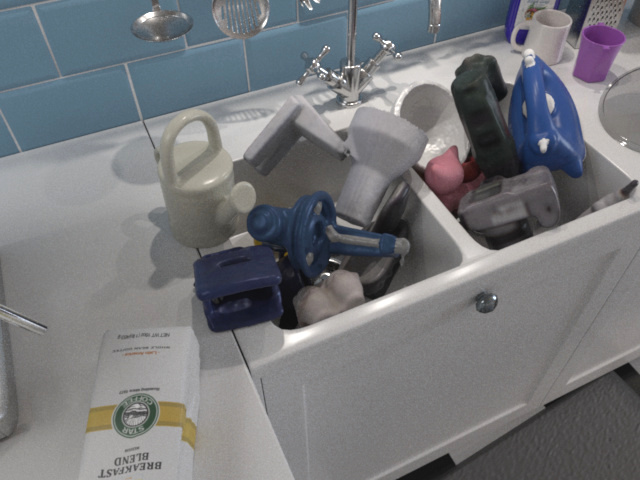} &
			\includegraphics[width=0.245\linewidth]{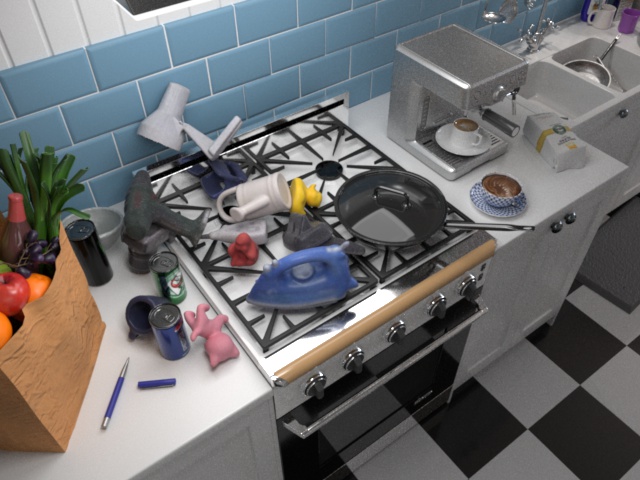} \\

			\includegraphics[width=0.245\linewidth]{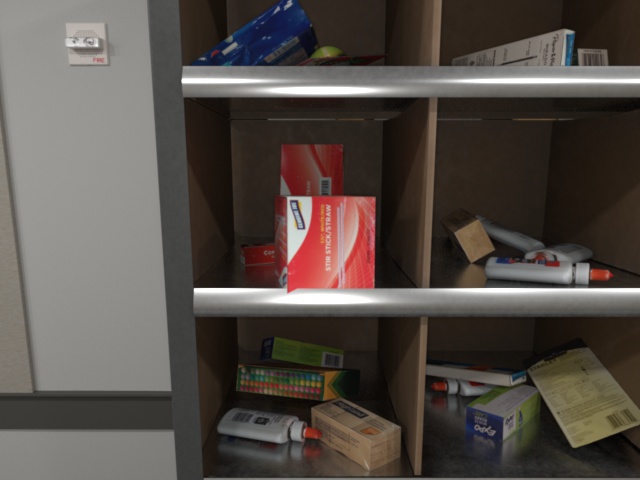} &
			\includegraphics[width=0.245\linewidth]{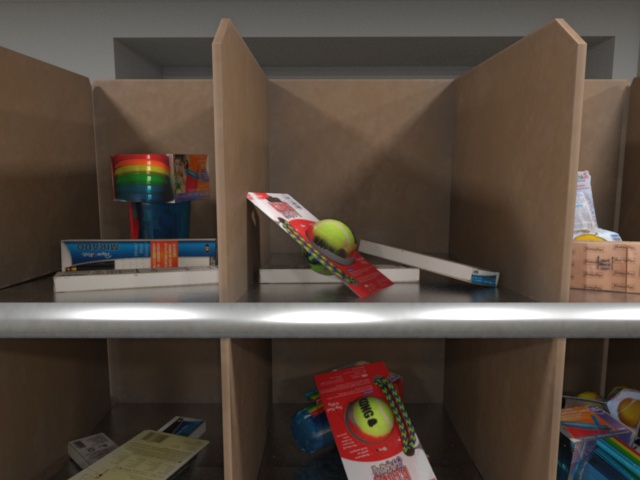} &
			\includegraphics[width=0.245\linewidth]{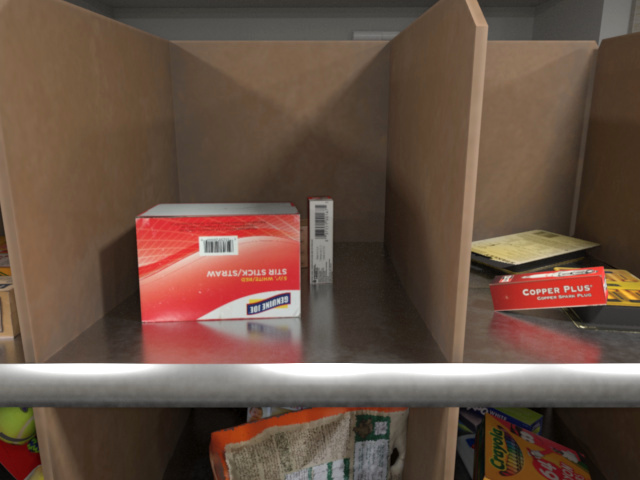} &
			\includegraphics[width=0.245\linewidth]{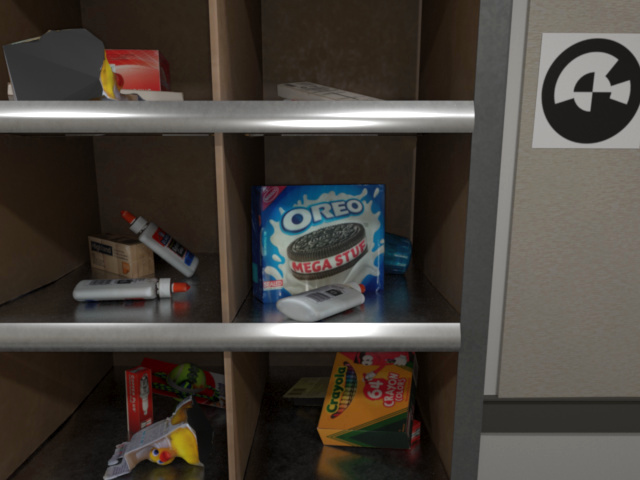} \\
		\end{tabular}
		\vspace{-2ex}
		\caption{\label{fig:supp_pbr_examples} Examples of high quality PBR images of objects from the LineMod dataset~\cite{hinterstoisser2012accv} in Scenes 1--5 (top five rows), and images of objects from the Rutgers APC dataset~\cite{rennie2016dataset} in Scene 6 (bottom row). The images were automatically annotated with 2D bounding boxes, masks and 6D poses of visible object instances. 
		}
	\end{center}
\end{figure*}

\begin{figure*}
	\begin{center}
		\begin{tabular}{ @{}c@{ } @{}c@{ } @{}c@{ } @{}c@{ } }
			\includegraphics[width=0.245\linewidth]{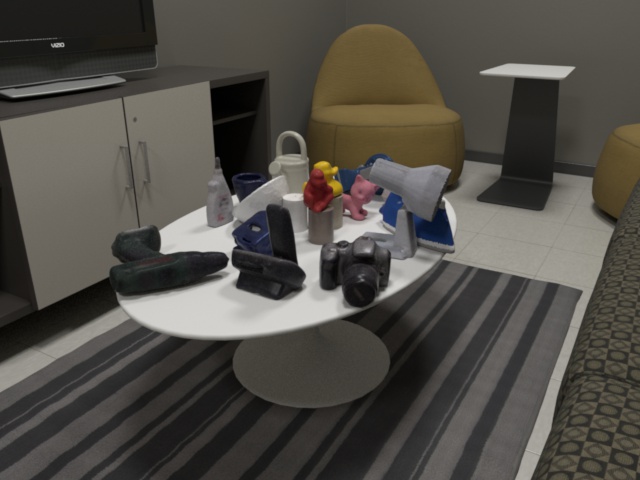} &
			\includegraphics[width=0.245\linewidth]{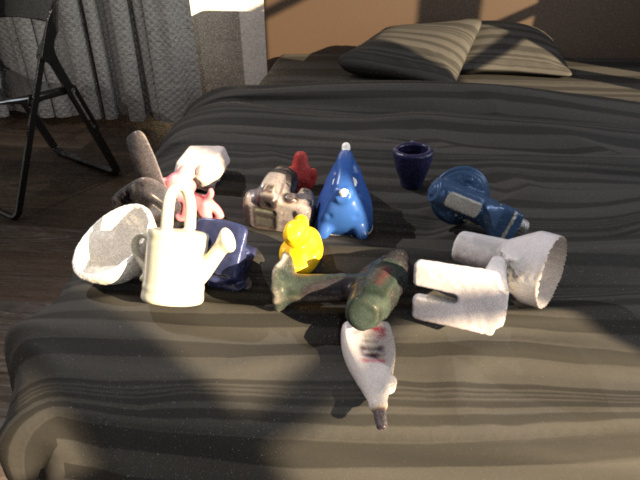} &
			\includegraphics[width=0.245\linewidth]{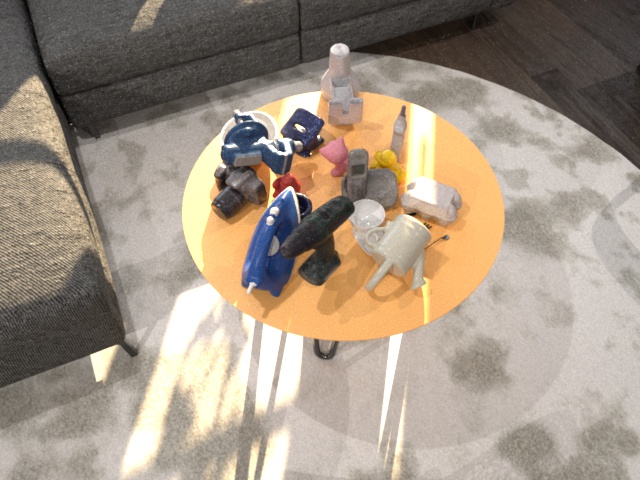} &
			\includegraphics[width=0.245\linewidth]{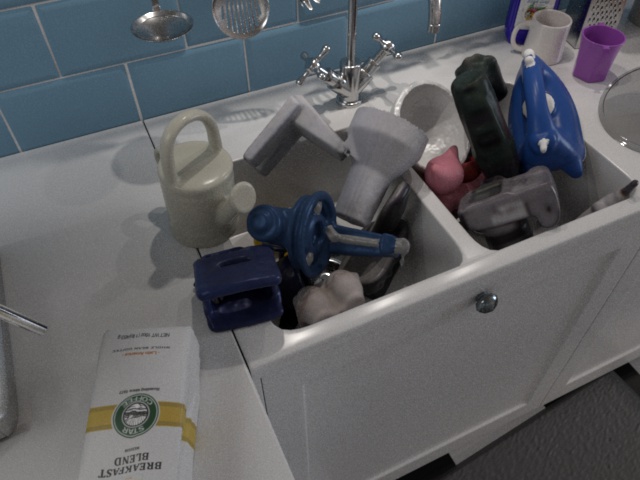} \\
			\includegraphics[width=0.245\linewidth]{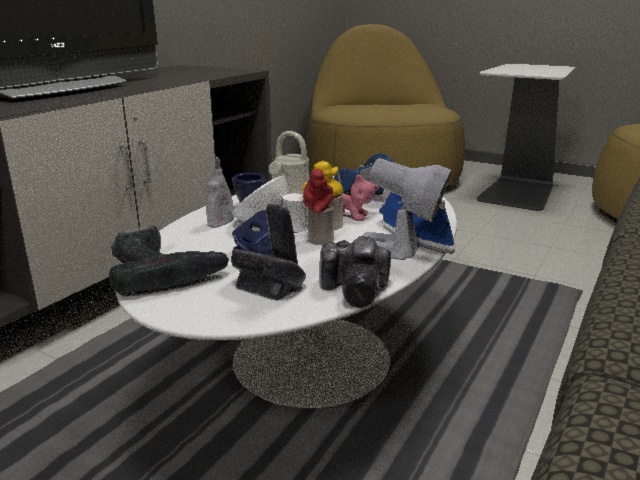} &
			\includegraphics[width=0.245\linewidth]{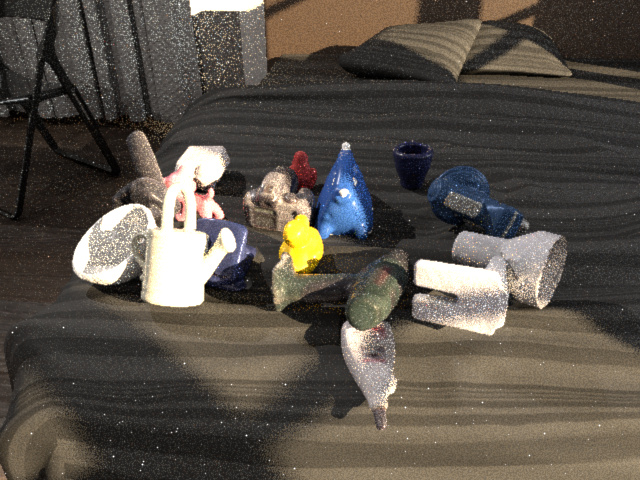} &
			\includegraphics[width=0.245\linewidth]{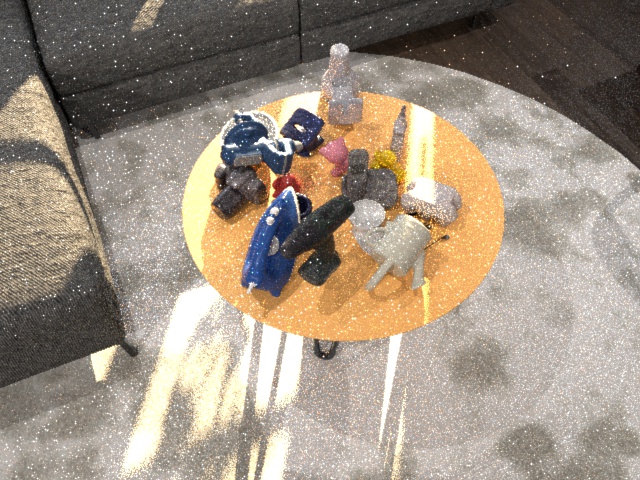} &
			\includegraphics[width=0.245\linewidth]{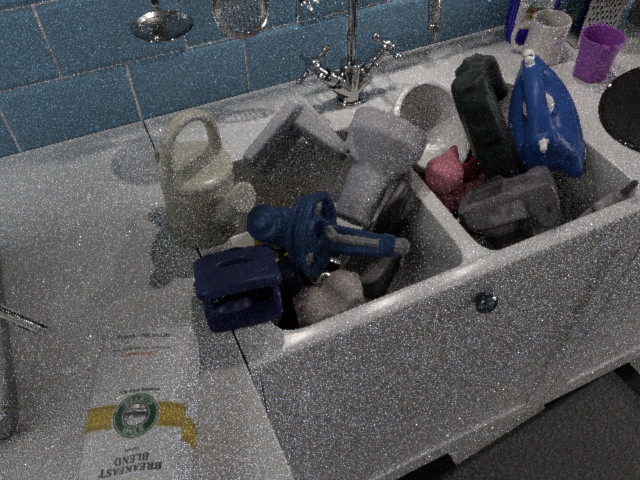} \\
		\end{tabular}
		\vspace{-2ex}
		\caption{\label{fig:supp_pbr_quality} The same images rendered in high (top) and low (bottom) PBR quality.}\vspace{-1ex}
	\end{center}
\end{figure*}

\begin{figure*}
	\begin{center}
		\begin{tabular}{ @{}c@{ } @{}c@{ } @{}c@{ } @{}c@{ } }
			\includegraphics[width=0.245\linewidth]{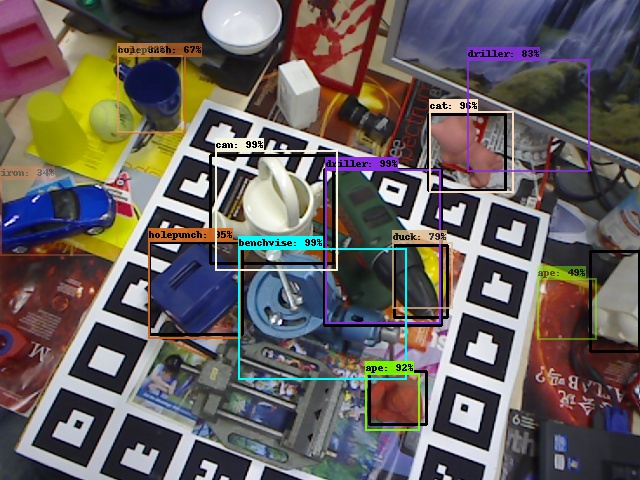} &
			\includegraphics[width=0.245\linewidth]{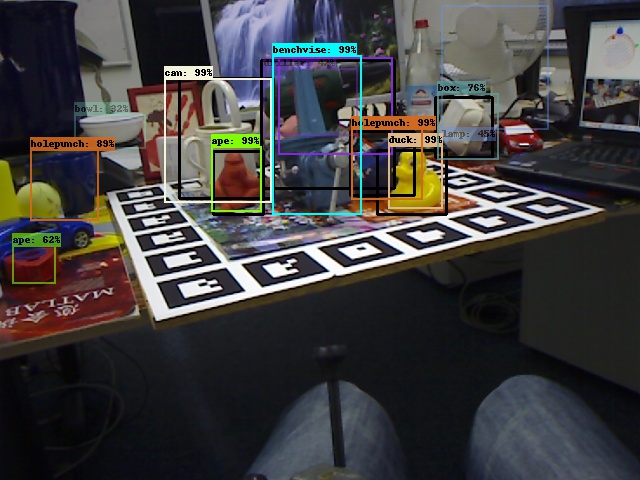} &
			\includegraphics[width=0.245\linewidth]{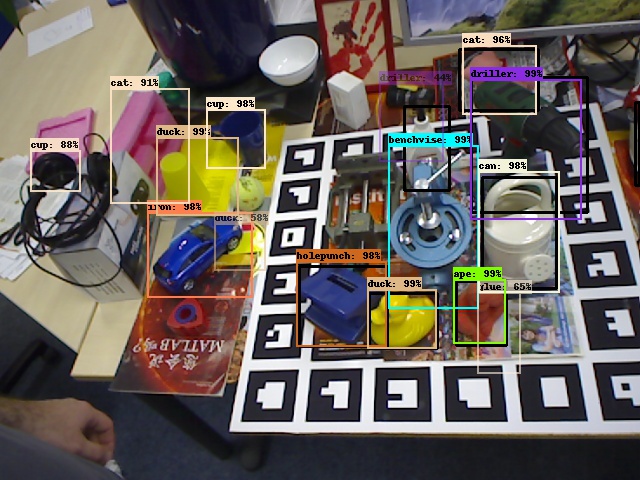} &
			\includegraphics[width=0.245\linewidth]{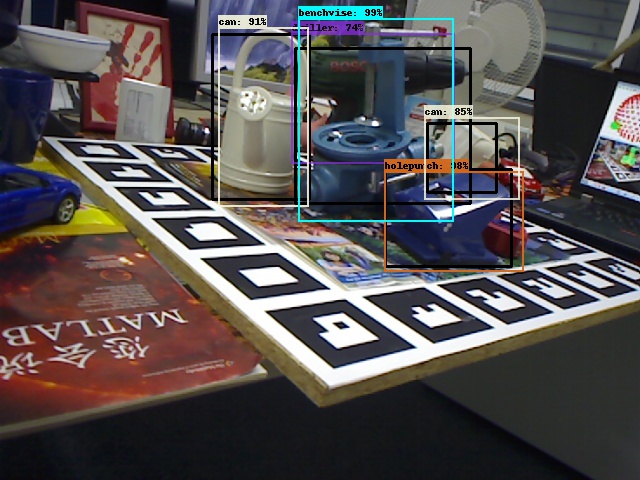} \\
			\includegraphics[width=0.245\linewidth]{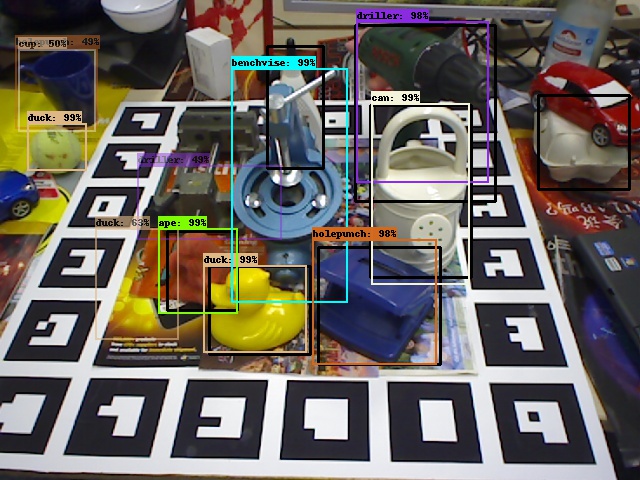} &
			\includegraphics[width=0.245\linewidth]{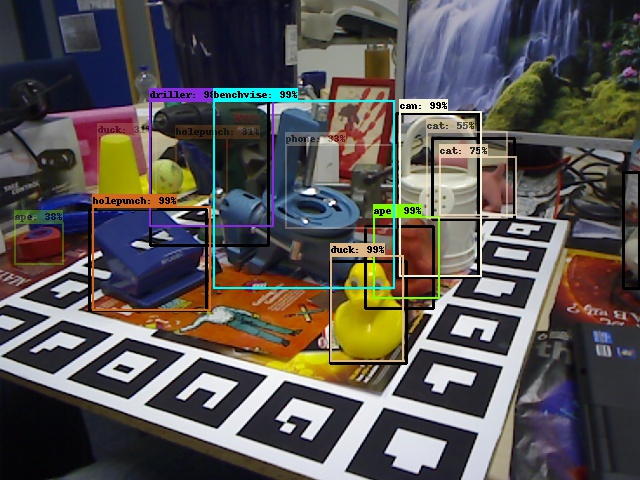} &
			\includegraphics[width=0.245\linewidth]{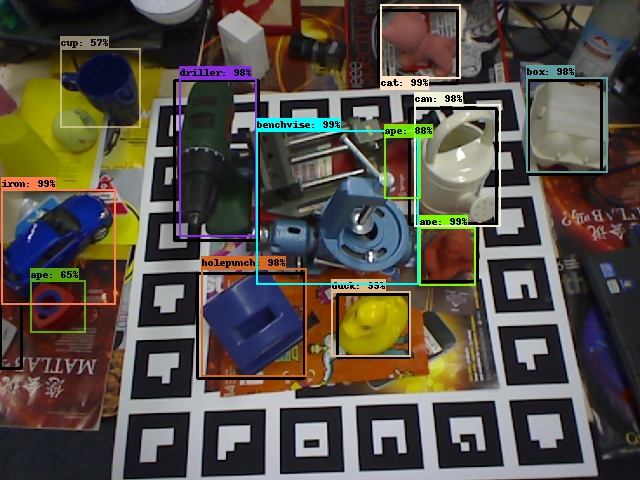} &
			\includegraphics[width=0.245\linewidth]{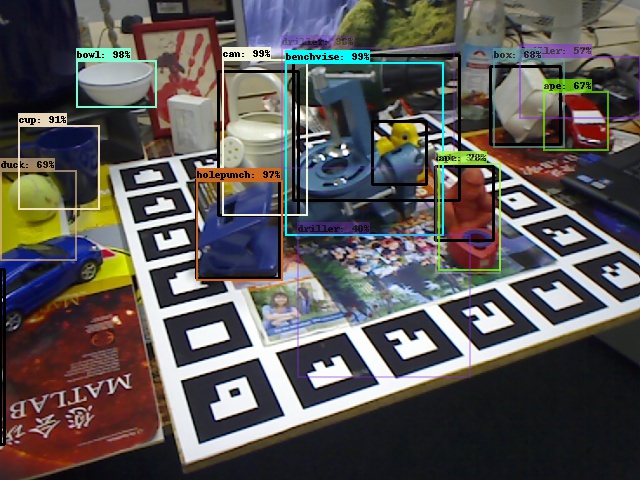} \\
			\includegraphics[width=0.245\linewidth]{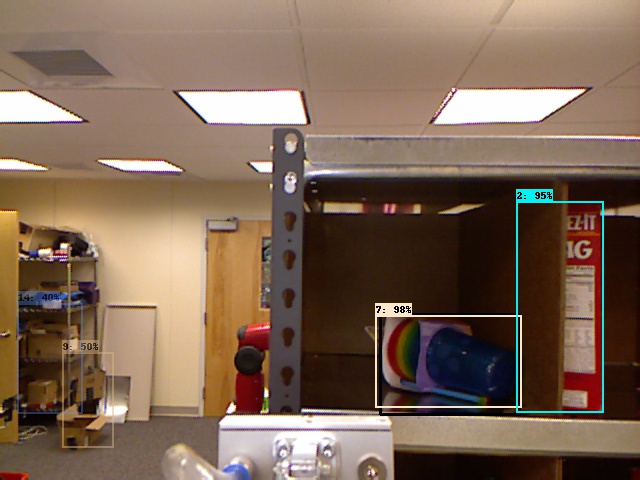} &
			\includegraphics[width=0.245\linewidth]{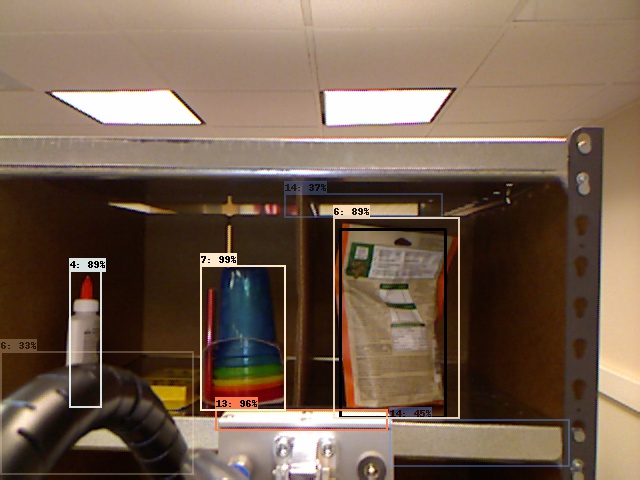} &
			\includegraphics[width=0.245\linewidth]{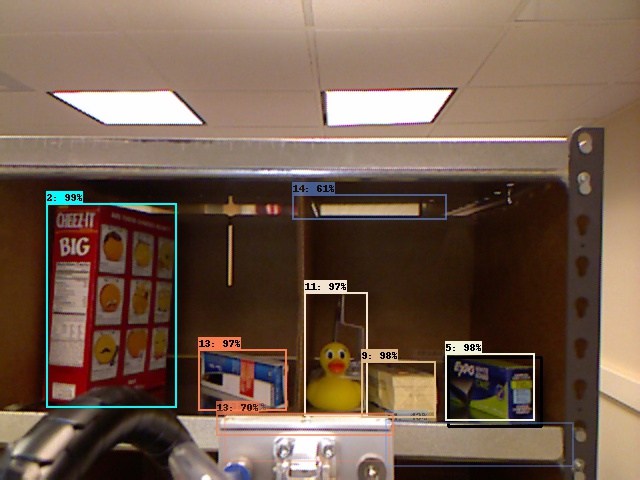} &
			\includegraphics[width=0.245\linewidth]{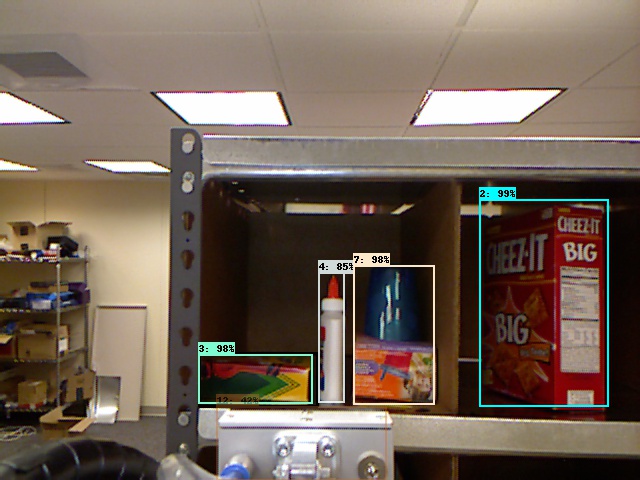} \\
			\includegraphics[width=0.245\linewidth]{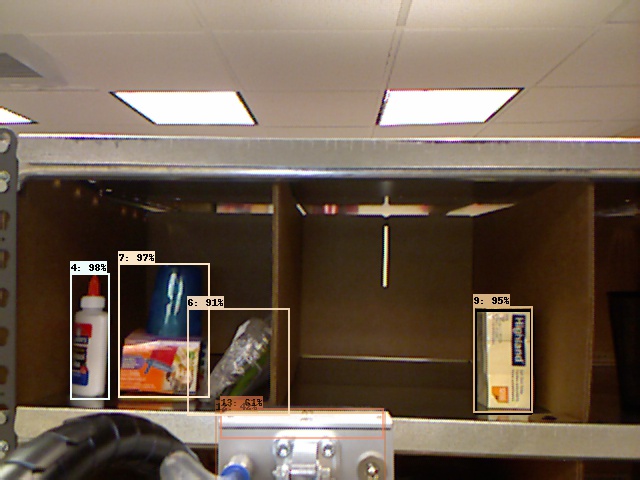} &
			\includegraphics[width=0.245\linewidth]{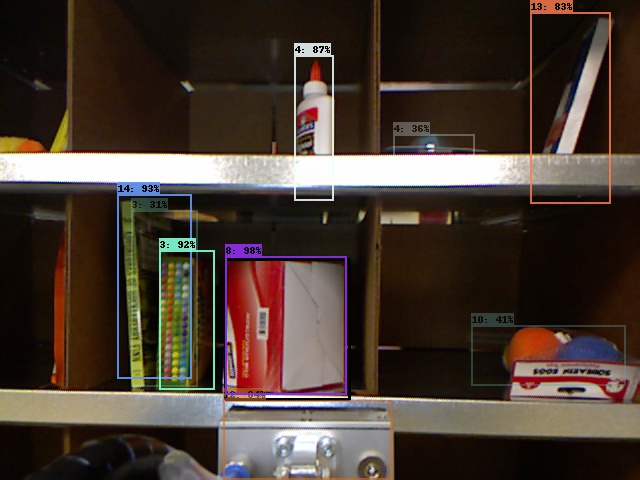} &
			\includegraphics[width=0.245\linewidth]{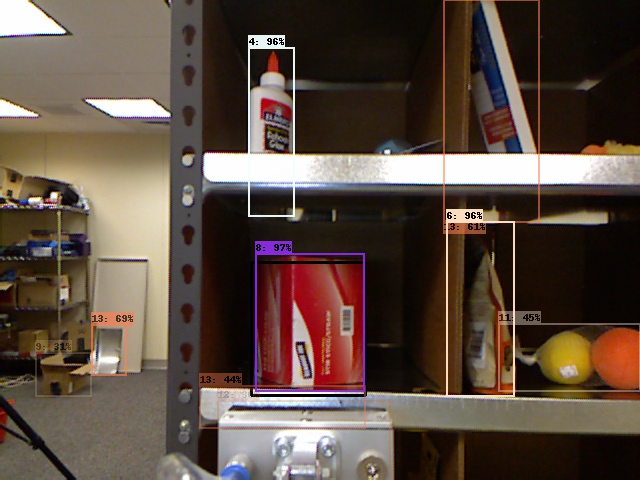} &
			\includegraphics[width=0.245\linewidth]{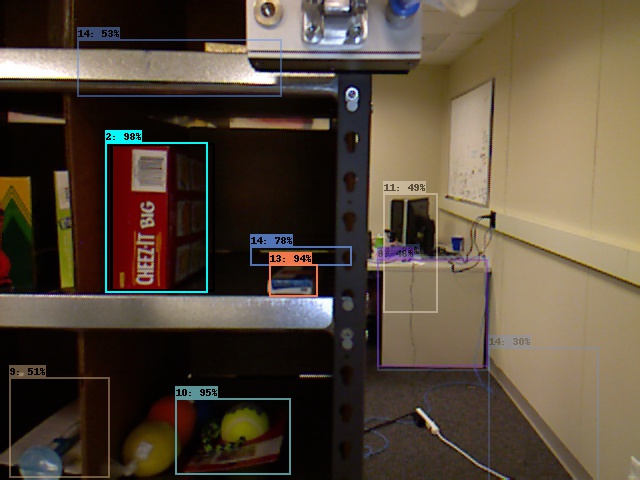} \\
		\end{tabular}
		\vspace{-2ex}
		\caption{\label{fig:supp_frcnn_results} Example results of the Faster R-CNN object detector~\cite{ren2017faster} trained on high quality PBR images and evaluated on real test images from the LineMod-Occluded dataset~\cite{BrachmannKMGSR_eccv14} (top two rows) and the Rutgers APC dataset~\cite{rennie2016dataset} (bottom two rows).
		}
	\end{center}
\end{figure*}

\noindent This supplement provides examples of the baseline training images in Fig.~\ref{fig:random_bg}, a visualization of the object pose generation process in Fig.~\ref{fig:arrangement}, images of out-of-context RU-APC objects in Fig.~\ref{fig:apc_out_of_context}, examples of high quality PBR images in Fig.~\ref{fig:supp_pbr_examples}, a comparison of low/high PBR quality in Fig.~\ref{fig:supp_pbr_quality}, example results of the Faster R-CNN object detector~\cite{ren2017faster} trained on high quality PBR images in Fig.~\ref{fig:supp_frcnn_results}, per-class detection scores in Tab.~\ref{tab:pbr-RU-APC} and Tab.~\ref{tab:pbr-LMO}, and a description of the PBR quality settings below.

\customparagraph{PBR Quality Settings.}
Tab.~\ref{fig:arnoldsampling} shows the mapping between the used quality settings and the Arnold parameters~\cite{georgiev2018arnold}. Increasing the number of rays traced per image pixel (\emph{AA}) reduces aliasing artifacts caused by insufficient sampling of geometry and noise caused by insufficient sampling of illumination. Increasing the number of diffuse rays traced when a ray hits a diffuse surface (\emph{D rays}) and the number of rays traced when the ray hits a specular surface (\emph{S rays}) reduces illumination noise. Increasing the number of diffuse and specular reflections (\emph{D depth} and \emph{S depth}) improves the accuracy of the illumination integral by gathering more of the light energy bounced around in the scene. This is especially important in the case when photons must bounce multiple times from the light source to reach an object visible to the camera.

\begin{figure}[h!]
	\scriptsize
	\begin{center}
		\begin{tabular}{|c|c|c|c|c|c|c|}
			\hline
			\textbf{Setting} & AA & D rays  & S rays & D depth  & S depth  & Max depth  \\
			\hline
			low & 1 & 1 & 1 & 1 & 1 & 2\\
			\hline
			medium & 9  & 36 & 36 & 3 & 2 &3 \\
			\hline
			high & 25 & 225 & 100 & 3 & 3 & 4 \\
			\hline
		\end{tabular}
	\end{center}
	\vspace{-2ex}
	\captionof{table}{\label{fig:arnoldsampling} Arnold parameters~\cite{georgiev2018arnold} for different quality settings.
	}
\end{figure}

	{\small
		\bibliographystyle{IEEEbib}
		\bibliography{ref}
	}

\begin{table*}[t!]
	\begin{center}
		\small
		\begin{tabular}{|c||c|c|c|c|c|c|c|c|c|c|c|c|c|c|c|}
			\hline
			Data/Obj. ID & 1 & 2 & 3 & 4 & 5 & 6 & 7 & 8 & 9 & 10 & 11 & 12 & 13 & 14 & mAP   \\ \hline									
			\hline
			\multicolumn{16}{|c|}{Inception-ResNet-v2}   \\ \hline
			PBR-h & 57.1 & 93.3 & \textbf{88.0}	& 61.2 & 80.4 & \textbf{62.5} & \textbf{99.0} & 98.1 & \textbf{73.2}	& \textbf{44.8} & 65.4 & \textbf{70.9} & 86.8 & \textbf{26.4} &	71.9 \\
			PBR-l & \textbf{57.6} & \textbf{96.3} & 84.6 & \textbf{62.2} & \textbf{81.3} & 60.5 & 98.7 & \textbf{98.6} & 73.1 & \textbf{44.8} & \textbf{79.3} & 67.2 & \textbf{90.5} & 25.6 & \textbf{72.9} \\
			PBR-ho & 46.2 & 61.9 & 56.0	& 55.8 & 54.4 & 69.8 & 89.0 & 89.3 & 81.6 & 21.5 & 72.7 &	58.3 & 43.3 & 21.7 & 58.7 \\
			BL & 33.5 & 47.5 & 71.5 & 32.7 & 42.4 & 13.5 & 44.9	& 73.0 & 57.4	& 44.5 & 47.6 & 35.8 & 87.6	& 40.6	& 48.0 \\ \hline \hline
			
			\multicolumn{16}{|c|}{ResNet-101}   \\ \hline
			PBR-h & 30.6 & \textbf{93.7} & \textbf{91.6} & \textbf{68.2} & 72.1 & 56.7 & 93.4 & \textbf{93.6} & \textbf{75.2} & \textbf{42.6} & \textbf{84.5} & 60.9 & 73.2 & 21.4 & \textbf{68.4} \\
			PBR-l & 26.8 & 87.6 & 87.3 & 64.0 & \textbf{79.8} & 27.8 & \textbf{95.2} & 90.4 & 66.2 & 37.5 & 83.1 & \textbf{61.4} & \textbf{79.3} & \textbf{25.3} & 65.1 \\
			PBR-ho & \textbf{35.2} & 64.4 & 58.4 & 52.9 & 46.7 & 53.0 & 71.5	& 73.8 & 69.3 & 32.2 & 66.2 & 51.8 & 28.3 &	19.3 & 51.6 \\
			BL & 29.1 & 38.5 & 82.0 & 59.2 & 52.4 & \textbf{59.1} & 79.5 & 75.0 & 36.4 & 36.8 & 75.1 & 50.6 & 48.5 & 14.8 & 52.7 \\
			\hline
		\end{tabular}
		\caption{\label{tab:pbr-RU-APC} \textbf{Object detection scores on RU-APC:} Per-class average precision (AP@.75IoU) and the mean average precision (mAP@.75IoU) of~Faster R-CNN trained on (i)~high/low quality PBR images of in-context objects rendered in Scene 6 (PBR-h, PBR-l), (ii)~high quality PBR images of out-of-context objects rendered in Scene 3 (PBR-ho), and (iii)~images of objects rendered on top of random photographs~(BL). The object identifiers follow the BOP convention~\cite{hodan2018bop}.}
	\end{center}
\end{table*}

\begin{table*}[ht!]
	\begin{center}
		\small
		\begin{tabular}{|c|c|c|c|c|c|c|c|c|c|}
			\hline									
			Data/Obj. ID & 1 & 5 & 6 & 8 & 9 & 10 & 11 & 12 & mAP   \\ \hline									
			\hline		
			\multicolumn{10}{|c|}{Inception-ResNet-v2}   \\ \hline
			PBR-h & \textbf{60.3} & \textbf{44.5} & \textbf{56.7} & \textbf{53.4} & \textbf{81.8} & \textbf{48.6} & 9.6 & \textbf{92.3} & \textbf{55.9} \\
			PBR-l & 57.3	& 35.8	& 53.3	& 52.6	& 77.8	& 23.8	& 3.1	& 94.5 & 49.8 \\
			BL & 30.7 & 45.4	& 42.5	& 32.4	& 77.1	& 33.4 & \textbf{19.6}	& 76.7	& 44.7 \\ \hline \hline								
			\multicolumn{10}{|c|}{ResNet-101}   \\ \hline	
			PBR-h & \textbf{46.3} & \textbf{40.3} & \textbf{48.5} & \textbf{58.0}	& \textbf{76.4} & \textbf{39.5} & 4.7	& \textbf{85.5}	& \textbf{49.9} \\
			PBR-l & 44.1	& 26.6	& 41.6	& 53.7	& 73.7	& 24.5	& 1.1	& 91.6	& 44.6 \\
			BL & 35.5 & 45.3 & 37.1 & 44.6 & 75.0 & 33.6 & \textbf{12.7} & 76.8 &	45.1 \\
			\hline
		\end{tabular}
		\caption{\label{tab:pbr-LMO} \textbf{Object detection scores on LM-O:}  Per-class average precision (AP@.75IoU) and the mean average precision (mAP@.75IoU) of~Faster R-CNN trained on (i) high/low PBR images of objects rendered in Scenes 1--5 (PBR-h, PBR-l), and (ii) images of objects rendered on top of random photographs (BL). The object identifiers follow the BOP convention~\cite{hodan2018bop}.}
	\end{center}
\end{table*}

\newpage
\null
\vfill